\documentclass[a4paper, 11pt]{article}
\usepackage[utf8]{inputenc}
\usepackage[margin=2cm]{geometry}

\usepackage{cite}
\usepackage{amsmath,amssymb,amsfonts}
\usepackage{algorithmic}
\usepackage{graphicx}
\usepackage{textcomp}

\usepackage{subcaption}
\usepackage{xcolor}
\usepackage{color, colortbl}
\usepackage{bbm}
\usepackage{multirow}
\usepackage{float}
\usepackage{arydshln}
\usepackage{pifont}
\usepackage{booktabs}
\usepackage{array}
\definecolor{Gray}{gray}{0.8}
\usepackage{relsize}
\usepackage{cite}
\usepackage{siunitx}
\usepackage{cleveref}
\usepackage{stmaryrd}
\usepackage{enumitem}
\usepackage{url}

\providecommand{\keywords}[1]{\textbf{\textit{Keywords---}} #1}

\newcommand{\eg}{\textit{e}.\textit{g}., }
\newcommand{\ie}{\textit{i}.\textit{e}.\ }

\makeatletter
\def\blfootnote{\xdef@thefnmark{}@footnotetext}
\makeatother

\usepackage[symbol]{footmisc}

\begin{document}

\begin{center}
\Large \textbf{Unpaired Translation from Semantic Label Maps to Images\\ by Leveraging Domain-Specific Simulations} \\
\vspace{.5em}
\small Lin Zhang$^{1}$, Tiziano Portenier$^1$, Orcun Goksel$^{1,2}$\\
\vspace{.5em}
\footnotesize
$^1$ Computer-assisted Applications in Medicine, Computer Vision Lab, ETH Zurich, Switzerland\\
$^2$ Department of Information Technology, Uppsala University, Sweden
\end{center}
\footnotetext{Funding was provided partially by the Medtech Science and Innovation Centre, Uppsala, Sweden.}
\footnotetext{Corresponding author: Orcun Goksel (orcun.goksel@it.uu.se)}

\begin{abstract}
Photorealistic image generation from simulated label maps are necessitated in several contexts, such as for medical training in virtual reality.
With conventional deep learning methods, this task requires images that are paired with semantic annotations, which typically are unavailable.
We introduce a contrastive learning framework for generating photorealistic images from simulated label maps, by learning from unpaired sets of both.
Due to potentially large scene differences between real images and label maps, existing unpaired image translation methods lead to artifacts of scene modification in synthesized images.
We utilize simulated images as surrogate targets for a contrastive loss, while ensuring consistency by utilizing features from a reverse translation network.
Our method enables bidirectional label-image translations, which is demonstrated in a variety of scenarios and datasets, including laparoscopy, ultrasound, and driving scenes.
By comparing with state-of-the-art unpaired translation methods, our proposed method is shown to generate realistic and scene-accurate translations.
\end{abstract}


\section{Introduction}
\label{sec:introduction}
Photorealistic image simulation has been an active research area for decades with a wide range of applications from movie- and game-industries~\cite{masuch2004game, tewari2020state} to medical imaging for surgical training~\cite{frangi2018simulation,elhelw2004real,burger2013real,mattausch2018realistic}.
Extensive research in modelling imaging physics~\cite{pharr2016physically,mattausch2018realistic,burger2013real} and material representations~\cite{weyrich2009principles,zhang2020deep} has substantially improved simulation realism in different applications, but there is still a very perceiveable visual difference between the state-of-the-art simulators and real world images.
Recent progress in deep learning has paved the way for synthesizing photorealistic images by learning image features from large-scale real data.
Among them, generative adversarial networks~\cite{goodfellow2014generative} have shown promising results in generating photorealistic images.
Methods were shown to successfully generate images from noise inputs with controlled styles at different level of details learned from given domains~\cite{karras2019style,karras2020analyzing, karras2021alias} as well as to achieve semantic image synthesis~\cite{isola2017image, park2019semantic,wang2018high,liu2019learning,jiang2020tsit} given paired label and target images.
In the absence of paired training samples, methods based on cycle consistency loss~\cite{zhu2017unpaired,yi2017dualgan}, shared latent space~\cite{huang2018multimodal, liu2017unsupervised}, layer normalizations~\cite{kim2019u} and more recently contrastive loss~\cite{park2020contrastive,zheng2021spatially} have been investigated with varying levels of success.

As opposed to the widely used cyclic approaches, contrastive learning (CL) based unpaired image translation methods focus on translating in single direction by relaxing the strong bijective assumption, which has achieved impressive results in various unpaired translation settings~\cite{park2020contrastive}.
Compared to image-level feature contrasting for unsupervised classification and segmentation, patch-level contrasting is employed in~\cite{park2020contrastive} which enforces structural similarity between images.
Various approaches have been accordingly proposed for bridging the appearance gap between synthetic and real images~\cite{shrivastava2017learning, vitale2020improving, hoffman2018cycada, sharan2021mutually}, also further improved by leveraging auxiliary simulation information such as simulated semantic maps~\cite{tomar2021content} and geometry buffers (G-buffers) generated during 3D computer graphics rendering~\cite{richter2022enhancing}.
A parallel line of work investigate photographic transfer~\cite{li2017universal,luan2017deep,huang2017arbitrary,li2019learning}, aiming at translating the appearance of reference images to simulated contents; however, such methods require lengthy and difficult-to-parametrize optimizations for each single target image.
All above work aim to improve the realism of existing, sub-realistic (\eg simulated) images and hence require the existence of preceding, complex simulation and rendering methods.

\begin{figure*}
\centering
\includegraphics[width=1\textwidth]{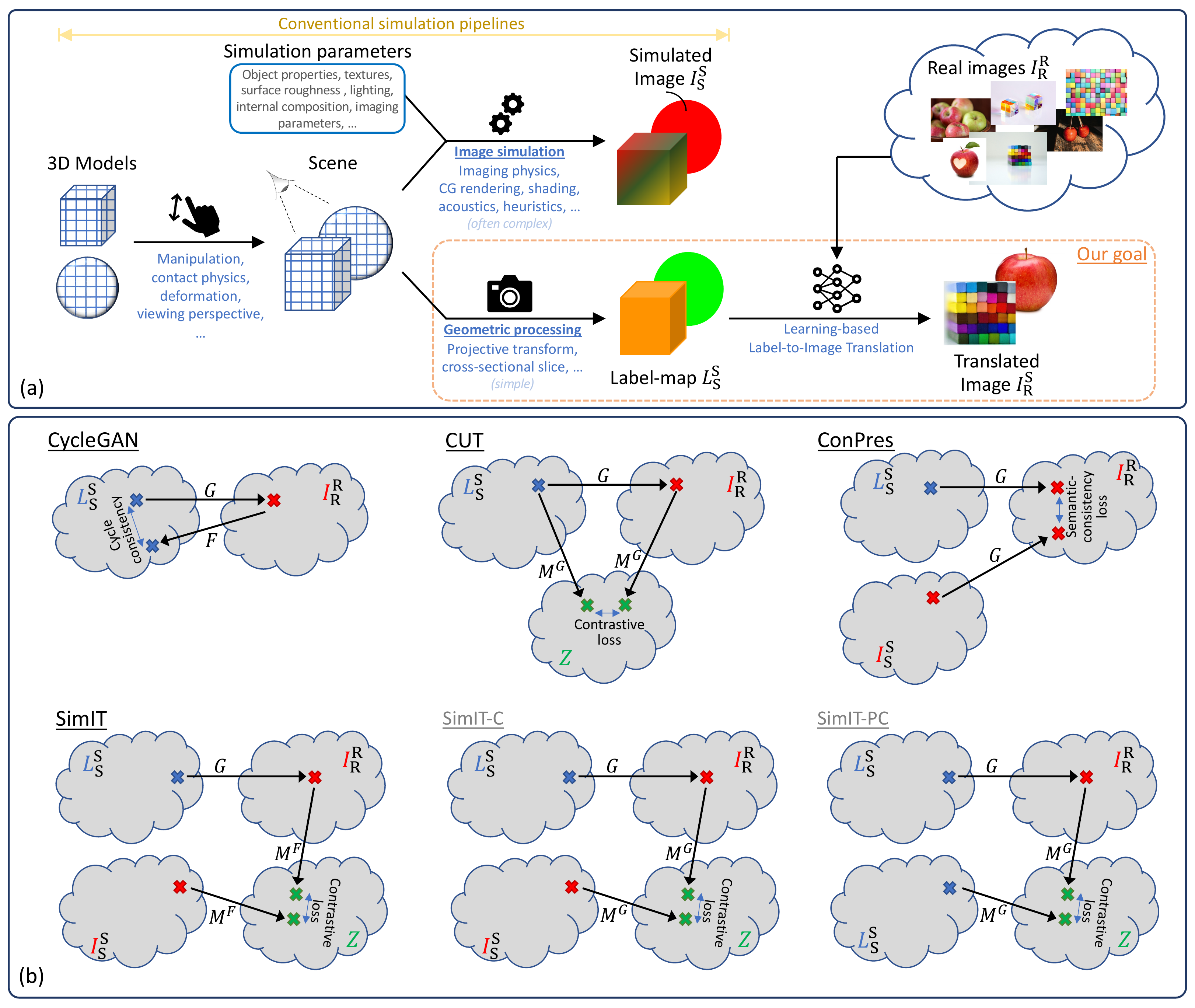}
\caption{Overview of unpaired label-image translation by leveraging domain-specific simulations. (a)~An illustration of the simulation/generation pipeline from 3D computer graphics (CG) model to label-maps $L^{\text{content}}_{\text{style}}$ and images $I^{\text{content}}_{\text{style}}$ with the subscript indicating the style domain and superscript indicating the content domain. $R$ and $S$ denote the real and simulated domains, respectively. 
Note that the goal is to generate images $I^\text{S}_\text{R}$ with realistic appearance for simulated content, based on (\ie consistent with) label-maps $L^\text{S}_\text{S}$ of simulated scenes.
To that end, one can collect and use many real-life images $I^\text{R}_\text{R}$, but these will not one-to-one match existing simulated content therefore preventing classical supervised training.
(b)~A schematic summary of existing unpaired image translation approaches CycleGAN~\cite{zhu2017unpaired}, CUT~\cite{park2020contrastive}, and ConPres~\cite{tomar2021content}, as well as our proposed methods SimIT, \mbox{SimIT-C}, and \mbox{SimIT-PC}. 
We define the label-to-image mapping function as $G: \mathbb{L}\rightarrow\mathbb{I}$ and the image-to-label mapping function as $F: \mathbb{I}\rightarrow\mathbb{L}$, with the label domain $\mathbb{L}$ and image domain $\mathbb{I}$.
Both $G$ and $F$ are parameterized with a deep neural network consisting of an encoder and a decoder, \ie $G(\cdot)=G^{\text{D}}(G^{\text{E}}(\cdot))$ and $F(\cdot)=F^{\text{D}}(F^{\text{E}}(\cdot))$.
Contrastive loss is computed on features obtained from the mappings $M^{\text{G}}(\cdot)=H(G^{\text{E}}(\cdot))$ or $M^{\text{F}}(\cdot)=H(F^{\text{E}}(\cdot))$, where $H$ maps the generator encoder latent to a feature space $\mathbb{Z}$.}
\label{intro}
\end{figure*}

Photorealistic image generation directly from simulated scene layouts, so-called \emph{label-maps}, would obviate any complex and computationally-intensive simulation/rendering process in real-time, by learning the internals of such rendering into a generative model during a non-time-critical offline learning stage.
Such label-maps can typically be extracted easily from existing simulation pipelines, only given 3D objects models and a vantage point (thus a scene), \ie  without a need to tune model-specific parameters nor to compute complex physical interactions.
To illustrate this further, a generic simulation pipeline is given in \Cref{intro}(a).
Given the above motivation, we aim to generate images $I$ with \emph{realistic} appearance but \emph{simulation}-controlled content, \ie $I^\text{S}_\text{R}$ as appearance and content of a representation are hereafter indicated in sub- and super-script, respectively.
With this convention, the methods from the literature mentioned earlier mostly target image-to-image translation of $I^\text{S}_\text{S}\rightarrow I^\text{S}_\text{R}$\,.
In comparison, label-to-image translation as we intend is often more challenging due to the large domain shift between these two representations.
Generating \emph{simulated} images from labels, \ie $L^\text{S}_\text{S}\rightarrow I^\text{S}_\text{S}$ translation, was studied in~\cite{zhang2020deep} for accelerating simulated image generation in real-time, for which a conditional GAN with supervised per-pixel loss was shown to provide promising results.
This, however, relatively simpler compared to our goal of generating $I^\text{S}_\text{R}$, since the former can be cast as a paired translation problem where the paired data ($L^\text{S}_\text{S},I^\text{S}_\text{S}$) is typically available from conventional simulation pipelines.
In contrast for our desired target of $I^\text{S}_\text{R}$, there exists no such paired label data.
A large domain gap together with the lack of paired data make our intended label-to-realistic-image translation very challenging, and, to the best of our knowledge, without any working solution so far.

In this work we target the above problem of photorealistic image generation directly from label-maps.
To facilitate the learning of appearance information from real images $I^\text{R}_\text{R}$\,, we propose to utilize any available physics-based simulation to generate intermediate image representations $I^\text{S}_\text{S}$\,.
We utilize these as a stepping stone to help bridge the domain gap between the labels $L^\text{S}_\text{S}$ and their real-image counterparts $I^\text{S}_\text{R}$ as desired.
To that end, we introduce a contrastive learning based image translation framework that leverages physics-based simulations/rendering in the training of unpaired label-to-image translation, but without needing such simulations during the real-time inference stage.
Compared to the existing works~\cite{richter2022enhancing,tomar2021content,zhang2020deep,shrivastava2017learning}, our proposed method performs image generation and realism enhancement simultaneously in a single step.
We demonstrate our method on enhancing medical image simulators for training, as well as car driving simulators for entertainment.

Our proposed solution builds on a bidirectional (cyclic) translation idea.
As a by-product of this design, it can also perform the inverse operation of image-to-label translation, \ie semantic image segmentation is also learned meanwhile in an unsupervised fashion without seeing any annotations of real images.
We also evaluate such segmentation outcomes in this work, as they opens future possibilities of alleviating annotation efforts.

\section{Results}\label{sec2}
\newcommand{\MM}{SimIT}

\subsection{Compared methods}
\vspace{1ex}\noindent{\bf Proposed method. }
We call our proposed method for generating realistic images from simulated semantic label-maps, with the target style learned from real images while retaining overall content matching the simulated scene, as \textbf{Sim}ulation-based \textbf{I}mage \textbf{T}ranslation framework (\MM).
Realistic and scene-accurate translation given unpaired data is herein enabled by two major contributions:
\begin{enumerate}[left=0pt .. \parindent]
    \item To address missing label-image pair information, we leverage existing physics-based simulations by using the simulated images (that are inherently paired with corresponding label-maps) as surrogate targets for contrastive learning.
    \item To enforce content/structure preservation, we devise a method that contrasts domain-specific image features extracted from a translation network that is trained using a cycle consistency loss.
    This further enables bidirectional translation, \ie in both label-to-image and image-to-label directions.
\end{enumerate}

\vspace{1ex}\noindent{\bf Compared methods. }
We evaluate SimIT comparatively to the following three state-of-the-art unpaired image translation methods:
\mbox{\bf CycleGAN}~\cite{zhu2017unpaired} is a conventional approach with cyclic consistency loss by employing separate generators and discriminators in each direction.
\mbox{\bf CUT}~\cite{park2020contrastive} is a unidirectional translation framework based on patch-based multi-scale contrastive loss computed on generator features.
\mbox{\bf ConPres}~\cite{tomar2021content} is a multi-domain translation framework that leverages simulated label-image pairs to retain structural content.
Together with cycle-consistency and contrastive losses, ConPres proposes a regularization loss for semantic consistency, which enforces a generator to create the same output for paired images and label-maps.
Consequently, \mbox{ConPres} can be used for both image-to-image and label-to-image translation.
The latter being the focus herein, we employ that use-case of \mbox{ConPres} in our comparisons.
High-level conceptual schematics of above-mentioned three approaches are illustrated in \Cref{intro}(b).

\vspace{1ex}\noindent{\bf Ablations. }
To further evaluate our two major contributions listed above, we ablated them cumulatively from SimIT, resulting in the following reduced models for comparison:
{\bf\MM-C} (SimIT without cycle loss) is a unidirectional version of SimIT, \ie without learning an inverse translation from image to labels, where the contrastive loss is then computed using features from the label-to-image generator, c.f.~SimIT-C in \Cref{intro}(b).
{\bf\MM-CS} (SimIT-C without leveraging simulations) does not utilize any simulation information, where the contrastive loss is then computed between semantic label-maps and translated images, similarly to \mbox{CUT} in \Cref{intro}(b).

\subsection{Evaluation}
All results are evaluated on unseen test data.
We employ the non-parametric two-sided Wilcoxon signed-rank test to assess differences between paired test results and report statistical significance with p-value.
Methodological and implementation details are given later in the Methods.
We compared the methods on three different applications (more details given in the Methods):

\vspace{1ex}\noindent{\bf Laparoscopy training. }
As physics-based simulation, computer-graphics rendering techniques were employed~\cite{harders2008surgical,tuchschmid2010high}, to simulate synthetic laparoscopic images from a 3D abdominal model.
During the rendering of each frame, a camera projection of anatomical labels provided the corresponding semantic label-maps.
This simulated dataset with paired image and label-maps are herein called as \emph{LaparoSim}.

For the laparoscopy application, we employed two different datasets of real images, thus evaluating two different target styles, called herein: 
\emph{Style-C}, represented by the public Cholec dataset containing 80 videos of cholecystectomy surgeries~\cite{endonet}; and 
\emph{Style-H}, represented by a single, in-house laparoscopic surgery video clip with a length of 13 minutes.
Sample images can be seen in \Cref{res_lapro}.

\vspace{1ex}\noindent{\bf Ultrasound training. }
The simulated images were generated using a ray-tracing framework~\cite{mattausch2018realistic} from a second trimester fetal model, by emulating a convex ultrasound probe at multiple locations and orientations on the abdominal surface, with imaging settings following~\cite{tomar2021content}.
Semantic label-map is rendered as a cross-section through the anatomical surfaces at the ultrasound center imaging plane.
We refer this simulated dataset as \emph{USSim}.

For the targeted real-image style, sample ultrasound images were collected using a GE Voluson E10 machine during standard fetal screening exams of 24 patients.
We refer this as \emph{GE-E10 style}.

\vspace{1ex}\noindent{\bf Gaming. }
As the gaming simulation, we used the \emph{GTA} dataset~\cite{richter2016playing} containing image-label pairs from a car-driving game.
For the real image style, we used the \emph{Cityscapes} dataset~\cite{Cordts2016Cityscapes} containing images of street scenes from German cities.

\subsection{Experiments}

\begin{figure*}
\centering
\includegraphics[width=0.95\textwidth]{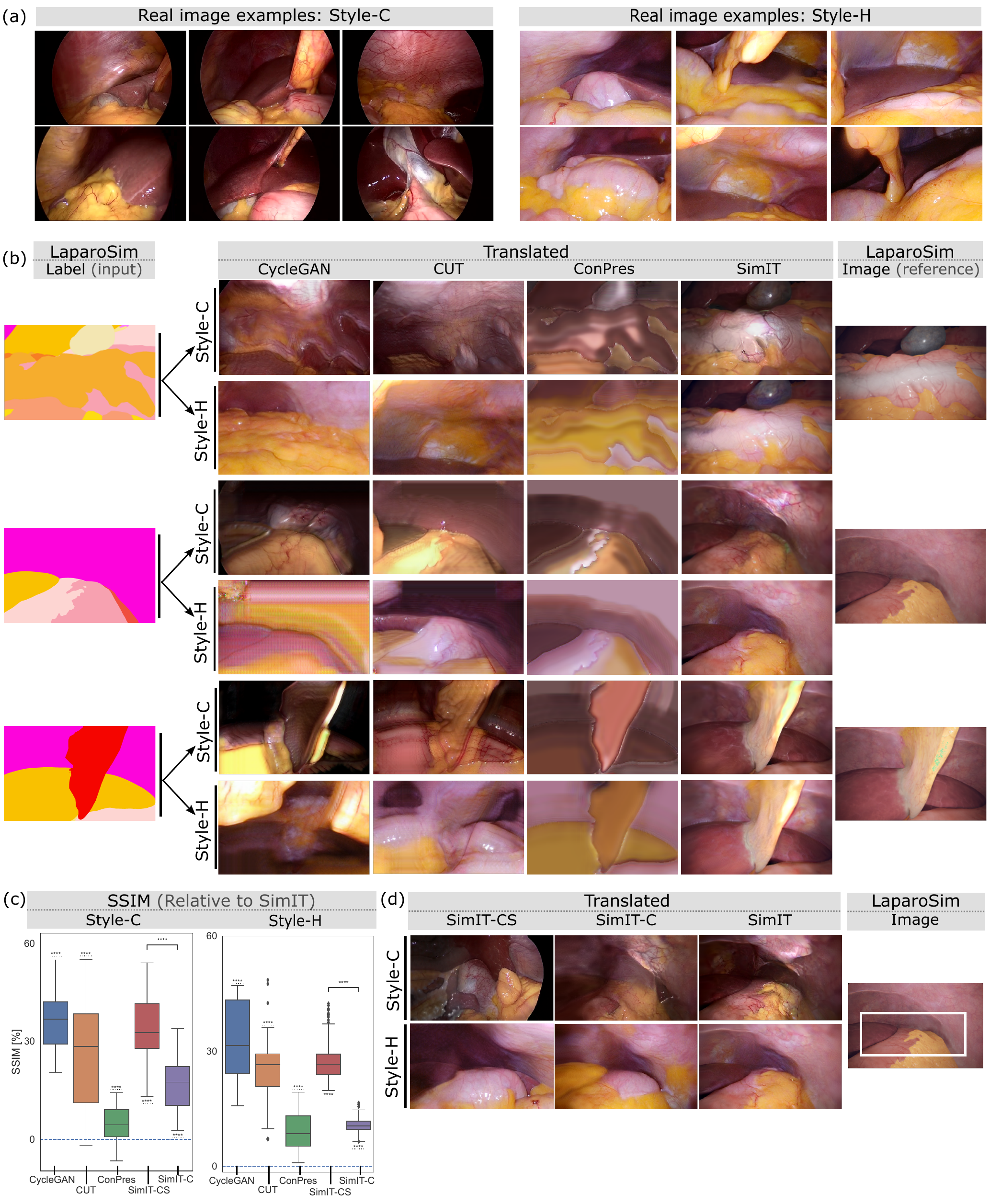}
\caption{(a)~Examples of real laparoscopic images with two different appearances: Style-C for the public Cholec80 dataset and Style-H for an in-house single-video dataset. 
(b)~Qualitative comparison of images translated from input LaparpSim label-maps, using the proposed SimIT and alternative methods. 
For reference purposes, conventionally simulated/rendered LaparoSim images are shown on the right.
(c)~Quantitative evaluation of structural preservation via Structure Similarity Index Metric (SSIM). Using a paired test, distributions of pair-wise differences over the test set are shown by comparing SimIT to each alternative method and ablated variant, \ie the larger the positive difference is, the more superior SimIT is with respect to another method. Significance is indicated with respect to SimIT (represented with the dotted lines) or between different models ($\mid$—$\mid$), with P-values of $\leq 0.0001$ (****).
(d)~Qualitative comparison of our proposed method SimIT to its ablated variants, with translated images zoomed in on the white field-of-view shown in the simulated image as reference.}
\label{res_lapro}
\end{figure*}

\vspace{1ex}\noindent{\bf Label-to-image translation. }
For the laparoscopy training application, we present the results in \Cref{res_lapro} for separately training two different styles.
As seen qualitatively in \Cref{res_lapro}(b), CycleGAN and CUT hallucinate inexistent tissue regions, \eg fat tissues.
ConPres achieves structural preservation by leveraging information from LaparoSim label-image pairs, but fails completely in generating tissue textures, which leads to highly unrealistic image style.
Going from label-to-image, our method SimIT is seen to outperform the state-of-the-art in terms of anatomical content preservation as well as in achieving a realistic image appearance.
This observation is substantiated by the quantitative evaluation in \Cref{quant}(a), where image realism is empirically measured using Frechet and Kernel Inception Distances (FID and KID, respectively) between translated and real image sets, and the content preservation is measured via the structural similarity index measure (SSIM) between translated and corresponding simulated images.
Note that SimIT also achieves the lowest SSIM standard deviation, indicating its consistent content preservation over the test samples.
A test-image wise paired comparison of all methods with respect to SimIT is presented in \Cref{res_lapro}(c), which shows ConPres as the closest contender in terms of content preservation (SSIM) but with largely unrealistic image translation, as also demonstrated qualitatively (\Cref{res_lapro}(b)) and tabulated empirically (\Cref{quant}(a)).

\newcommand{\std}[1]{{\relsize{-1}(#1)}}
\begin{table*}
\caption{Quantitative metrics reported as \emph{mean\std{std}}. Arrows indicate direction of superiority; $\uparrow$ meaning the higher the better, and $\downarrow$ the lower. KID is reported in $10^{-2}$ unit. Best results are marked bold.}
\begin{subtable}{1\textwidth}
    \centering
\begin{tabular}{c|l||ccc|ccc}
\toprule 
\multicolumn{2}{@{}l||}{\multirow{2}{*}{(a)~{\bf Laparoscopy}}} 
& \multicolumn{3}{c|}{Style-C} & \multicolumn{3}{c}{Style-H} \\
\cmidrule(lr){3-5} \cmidrule(lr){6-8}
 & \multirow{2}{*}{} & \multicolumn{1}{c}{Content} & \multicolumn{2}{c|}{Realism} & \multicolumn{1}{c}{Content} & \multicolumn{2}{c}{Realism} \\

 &\multicolumn{1}{c||}{\cellcolor[HTML]{C0C0C0}Method}
 &\multicolumn{1}{c}{\cellcolor[HTML]{C0C0C0}SSIM [$\%$] $\uparrow$}
  &\multicolumn{1}{c}{\cellcolor[HTML]{C0C0C0}KID $\downarrow$}
   &\multicolumn{1}{c|}{\cellcolor[HTML]{C0C0C0}FID $\downarrow$}
 &\multicolumn{1}{c}{\cellcolor[HTML]{C0C0C0}SSIM [$\%$] $\uparrow$}
  &\multicolumn{1}{c}{\cellcolor[HTML]{C0C0C0}KID $\downarrow$}
   &\multicolumn{1}{c}{\cellcolor[HTML]{C0C0C0}FID $\downarrow$} \\
\hline
 & Simulation  & --- &  257.39 &17.76 & ---  &201.32 &12.42 \\
\hdashline
\parbox[t]{2mm}{\multirow{3}{*}{\rotatebox[origin=c]{90}{Others}}}
& CycleGAN~\cite{zhu2017unpaired}  & 39.21\std{6.80}  &254.61 &14.46  & 50.50\std{10.62}  &212.42 &13.73 \\
& CUT~\cite{park2020contrastive}   & 49.79\std{13.75}  &234.65 &12.85  & 58.74\std{6.77}  &222.81 &13.42 \\
& ConPres~\cite{tomar2021content}   & 71.12\std{3.96}  &380.70 &36.72  & 75.76\std{5.56}  &379.82 &36.80 \\
\hdashline
\parbox[t]{2mm}{\multirow{3}{*}{\rotatebox[origin=c]{90}{Ours}}}
& SimIT-CS  & 41.77\std{7.98}  &\bf{202.24} &\bf{10.40}  & 56.15\std{5.23}  &\bf{147.06} &\bf{7.03} \\
& SimIT-C   & 58.05\std{7.34}  &210.94 &12.65   & 72.87\std{2.12}  &175.38 &11.61 \\
& SimIT  & {\bf 75}.{\bf 56}\std{2.42}  &214.22 &11.97   & \bf{83.69}\std{1.63}  &161.29 &7.13 \\
\bottomrule
\end{tabular}
\end{subtable}

\vspace{1ex}

\begin{subtable}{\textwidth}
  \centering
\begin{tabular}{c|l||c|c|c} 
\toprule 
\multicolumn{2}{@{}l||}{(b)~{\bf Ultrasound}} 
& \multicolumn{1}{c|}{Content} & \multicolumn{2}{c}{Realism} \\

 &\multicolumn{1}{c||}{\cellcolor[HTML]{C0C0C0}Method}
 &\multicolumn{1}{c|}{\cellcolor[HTML]{C0C0C0}IoU [$\%$] $\uparrow$}
 &\multicolumn{1}{c|}{\cellcolor[HTML]{C0C0C0}FID $\downarrow$}
 &\multicolumn{1}{c}{\cellcolor[HTML]{C0C0C0}KID  $\downarrow$}\\
\hline
 &Simulation & ---  &297.24 &38.14 \\
\hdashline
\parbox[t]{2mm}{\multirow{3}{*}{\rotatebox[origin=c]{90}{Others}}}
 &CycleGAN~\cite{zhu2017unpaired} & 2.33\std{1.10}  &46.67 &2.02  \\
 &CUT~\cite{park2020contrastive} & 3.39\std{1.47}  &\bf{46.62} &\bf{1.63}  \\
 &ConPres~\cite{tomar2021content} & 5.01\std{1.95}  &95.31 &7.74\\
\hdashline
\parbox[t]{2mm}{\multirow{3}{*}{\rotatebox[origin=c]{90}{Ours}}}
 &SimIT-CS  & 2.97\std{1.30}  &56.70 &3.57  \\
 &SimIT-C  & 9.50\std{4.42}  &46.10 &2.33  \\
  &SimIT & \bf{20.54}\std{6.80}  &79.02 &6.14 \\
  \bottomrule
\end{tabular}
\end{subtable}

\vspace{1ex}

\begin{subtable}{1\textwidth}
\centering
\begin{tabular}{c|l||c|c|c|c|c} 
\toprule 
\multicolumn{2}{@{}l||}{(c)~{\bf Gaming}} 
& \multicolumn{3}{c|}{Content} & \multicolumn{2}{c}{Realism} \\

 &\multicolumn{1}{c||}{\cellcolor[HTML]{C0C0C0}Method}
 &\multicolumn{1}{c|}{\cellcolor[HTML]{C0C0C0}mAP $\uparrow$}
 &\multicolumn{1}{c|}{\cellcolor[HTML]{C0C0C0}pixAcc $\uparrow$}
  &\multicolumn{1}{c|}{\cellcolor[HTML]{C0C0C0}classAcc $\uparrow$}
 &\multicolumn{1}{c|}{\cellcolor[HTML]{C0C0C0}FID $\downarrow$}
 &\multicolumn{1}{c}{\cellcolor[HTML]{C0C0C0}KID $\downarrow$}\\
\hline
 &Simulation & --  & -- & --  &101.82 &7.78 \\
\hdashline
\parbox[t]{2mm}{\multirow{3}{*}{\rotatebox[origin=c]{90}{Others}}}
 &CycleGAN~\cite{zhu2017unpaired} &14.12  &55.54 &24.40 &65.20 &3.04  \\
 &CUT~\cite{park2020contrastive} &12.42   &53.10 &22.06 &66.54 &3.42  \\
 &ConPres~\cite{tomar2021content} &15.44   &60.33 &26.38 &120.63 &10.39\\
\hdashline
\parbox[t]{2mm}{\multirow{3}{*}{\rotatebox[origin=c]{90}{Ours}}}
 &SimIT-CS  &13.74  &56.22 &23.19 &\bf{43.04} &\bf{0.81}  \\
 &SimIT-C  &20.76  &63.12 &32.57 &57.05 &1.89 \\
  &SimIT  &\bf{22.34}   &\bf{68.83}  &\bf{33.07} &60.46 &2.60 \\
  \bottomrule
\end{tabular}
\end{subtable}
\label{quant}
\end{table*}

Compared to the proposed method SimIT, its ablated variants, \mbox{SimIT-C} and \mbox{SimIT-CS} perform substantially poorer as seen quantitatively in \Cref{quant}(a) and \Cref{res_lapro}(c), and qualitatively in \Cref{res_lapro}(d).
This demonstrates the importance of our proposed method components.
\mbox{SimIT-CS} lacks our proposed component for utilizing simulations with a contrastive loss in learning the label-to-image translation, and as such it can be seen as a variant of CUT implemented in our framework.
With no explicit label-to-image pairs provided, \mbox{SimIT-CS} then learns to simply emulate all structures seen in the real examples, hence erroneously changing the image content as seen in the presented examples. 
Using simulated images as surrogate targets for contrastive loss (\mbox{SimIT-C} in \Cref{res_lapro}(d)) largely prevents such superfluous content generation.
Still \mbox{SimIT-C} only uses the features from a label domain for contrasting, and such features cannot be well aligned with image features.
With the proposed method \mbox{SimIT}, the addition of a custom cycle loss allows for training a bidirectional translation, where the features from an encoder operating on images can then instead be used for contrasting.
With such domain-consistent features, content preservation is further enhanced, as seen both quantitatively given the error of \mbox{SimIT-C} in \Cref{res_lapro}(c), and qualitatively by visually comparing these variants in \Cref{res_lapro}(d).

\begin{figure*}
\centering
\includegraphics[width=1\textwidth]{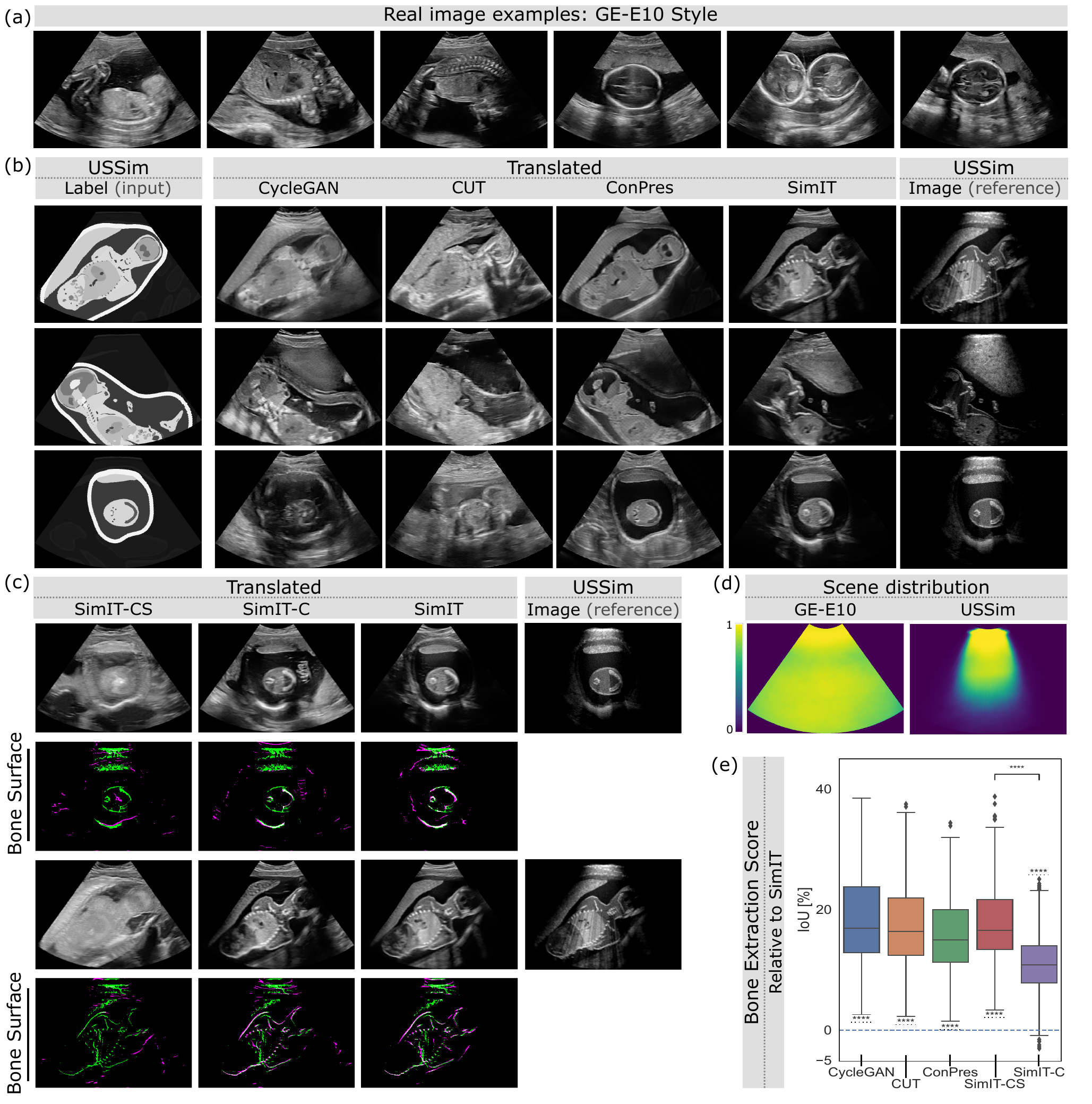}
\caption{Ultrasound training experiment results:
(a)~Examples of real ultrasound images with the style of GE-E10 ultrasound machine. 
(b)~Visual examples of images translated using SimIT compared to alternative methods. 
(c)~Qualitative results of SimIT compared to its ablated variants, with bone surfaces visualized in purple and green respectively for those from translated and simulated (reference) images. 
(d)~Probability maps of containing content at each location, averaged over the training images. A big scene distribution difference can be observed between simulated and target images. 
(e)~Quantitative evaluation of structural preservation. To this end, a paired test is employed to compare the IoU scores between the bone maps extracted from the simulated and translated images.
The difference is computed by subtracting the score of other models from SimIT, \ie the larger the positive difference is, the more superior SimIT is with respect to another method. Significance is indicated with respect to our proposed model SimIT (marked with dotted lines) or between different models ($\mid$—$\mid$), with P-values of $\leq 0.0001$ (****).}
\label{res_us}
\end{figure*}

\begin{figure*}
\centering
\includegraphics[width=1\textwidth]{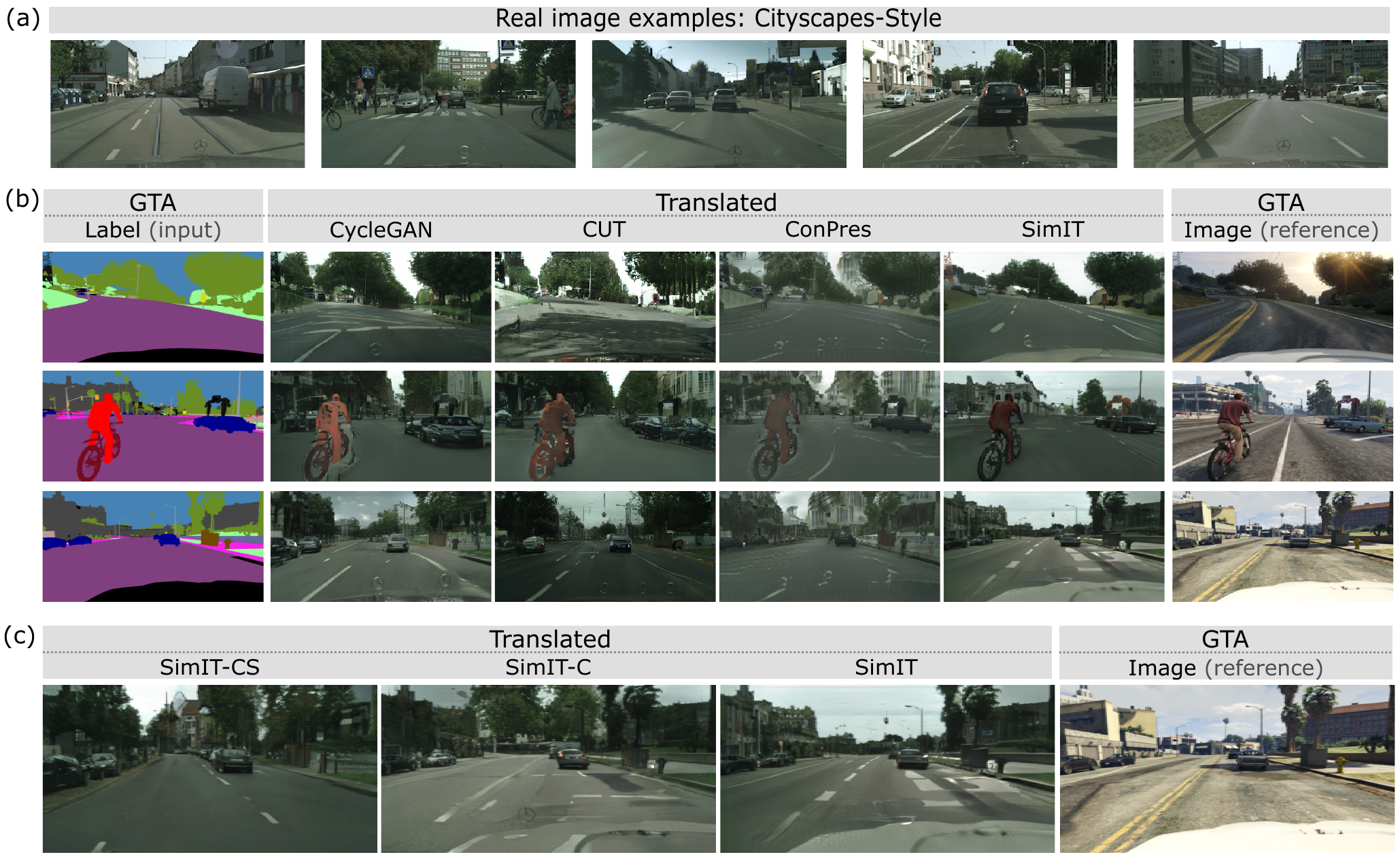}
\caption{Gaming experiment results: (a)~Examples of images from Cityscapes dataset. 
(b)~Visual examples of images translated using SimIT compared to alternative methods. 
(c)~Qualitative results of SimIT compared to its ablated variants.}
\label{res_gta}
\end{figure*}

Evaluation on ultrasound training and gaming applications further confirms the superior performance of our proposed method on label-to-image translation task (\Cref{res_us,res_gta}).
Translated ultrasound image examples in \Cref{res_us}(b) demonstrate that the alternative methods are not always correct with the echogenecity (brightness) profile of different anatomical regions, \eg outside the uterus is sometimes black and sometimes white, and the same for the amniotic fluid.
ConPres preserves the echogenecity better than CycleGAN and CUT by leveraging simulated label-image pairs, however, it is biased towards interpreting input labels as pixel intensities.
In comparison, SimIT can retain correct echogenecity of each region, which can be seen by comparing to the reference simulated images, while translating into a realistic appearance in the style of GE-E10 images.
Furthermore, our method successfully preserves fine anatomical structures, \eg the ribs in the top row example, whereas the other compared methods fail with such detail.
We herein assess content preservation for ultrasound images based on the alignment of bone surfaces, delineated by a phase-symmetry based bone-surface estimation method~\cite{hacihaliloglu2009bone} as exemplified in \Cref{res_us}(c).
Alignment is then quantified by the intersection over union (IoU) score of bone surface pixels extracted from translated and simulated (reference) images (cf.\ \Cref{res_us}(e) and \Cref{quant}(b)). 
When comparing image realism via FID/KID scores, SimIT does not yield the best values, which is hypothetically caused by the large scene difference between real and simulated training images, as illustrated in \Cref{res_us}(d).
Since the alternative methods do not enforce strict restrictions on content preservation, they hallucinate content in an unconstrained manner, which helps lower their FID/KID scores; nevertheless, such arbitrary content does not match the input label-maps, hence not fit for our purposes, as also quantified by the structural preservation scores, \ie IoU for the ultrasound training experiment.

For the gaming experiment, similarly to ultrasound experiment above, CUT and CycleGAN largely hallucinate content, ignoring the input GTA label-maps and hence do not satisfy the desired content preservation criterion.
For example, as seen in \Cref{res_gta}(b), the sky is erroneously filled with trees, since the real Cityscapes images contain more trees and less sky compared to the simulated GTA dataset~\cite{richter2022enhancing}.
Then, the discriminator can easily differentiate between real and fake images by looking at the image top region, which in return encourages the generator to hallucinate more trees in the sky.
In comparison to CycleGAN and CUT, the domain-consistent deep features used in our proposed SimIT explicitly encourages content preservation.
To evaluate structural consistency, we apply on the translated images the pretrained semantic segmentation network DRN~\cite{yu2017dilated} following~\cite{park2020contrastive} and report the resulting segmentation metrics: mean average precision (mAP), pixel accuracy (pixAcc), class accuracy (classAcc) -- see the Methods for details.
SimIT achieves the best scores among all the methods for these content preservation metrics.
Evaluating image realism using FID/KID scores (\Cref{quant}(c)), SimIT outperforms the state-of-the-art, while faring suboptimal compared to its ablated variants, which however in turn fail at successfully retaining content.
This well indicates the conflicting nature between content preservation and image realism, especially in the presence of substantial layout differences between simulated and real image sets.

\vspace{1ex}\noindent{\bf Image-to-label translation. }
By introducing the cyclic loss to train our framework with domain-consistent features, at the end of the training we also obtain a generator that can translate real images to label-maps, \ie a semantic segmenter for real images.
Note that such segmenter is trained truly unsupervised, \ie without requiring \emph{any} annotation of any real image.
To evaluate the segmentation outcome of image-to-label translations of \MM, we compare resulting label-maps to semantic segmentations for the datasets where such annotations are available, \ie the CholecSeg8k dataset as the Style-C target for our laparoscopy application and the Cityscapes dataset for our gaming application.

\begin{figure*}
\centering
\includegraphics[width=1\textwidth]{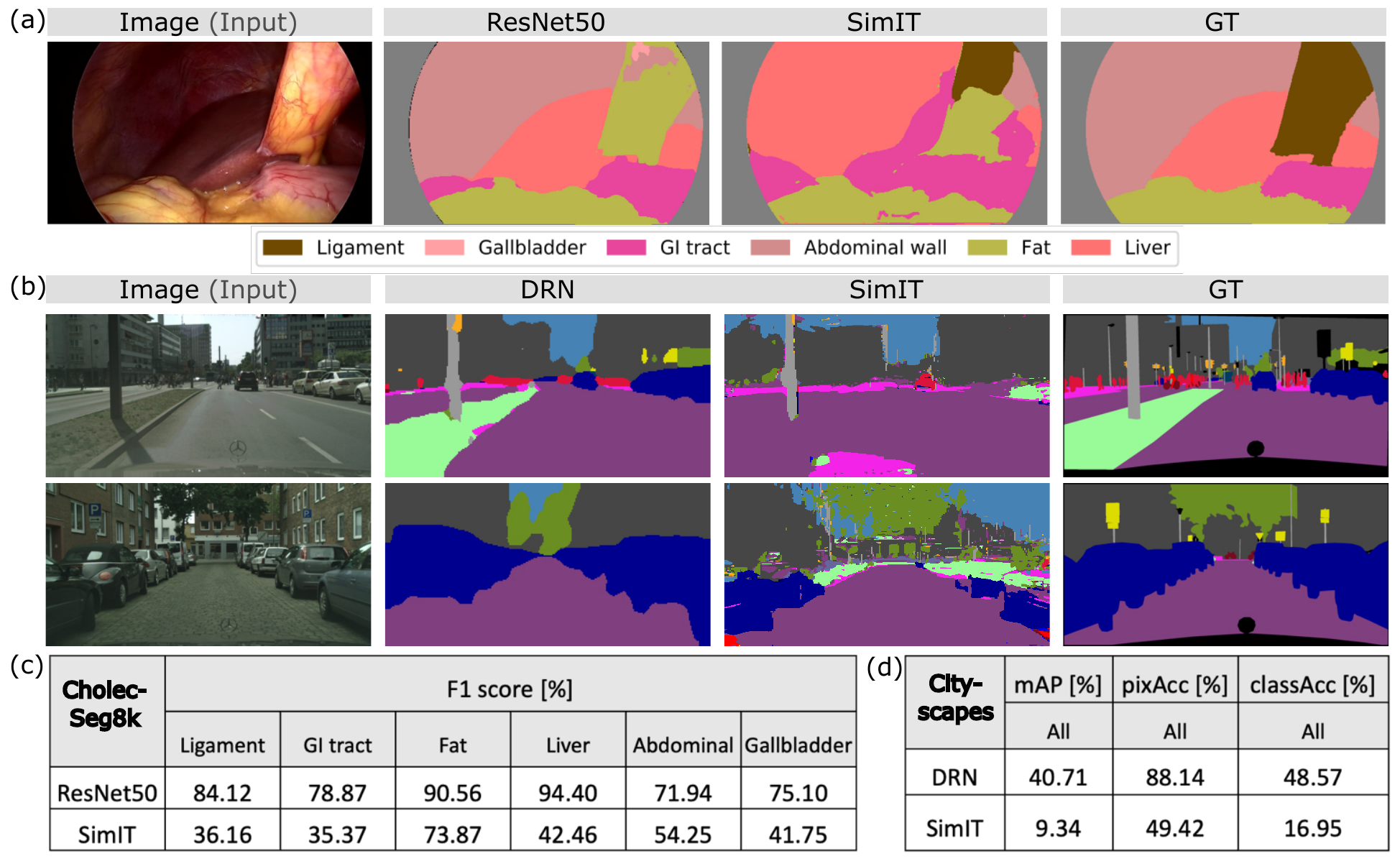}
\caption{(a)~A visual example of semantic label-maps predicted by SimIT compared with the supervised baseline ResNet50, with the label legend shown on the bottom. (b)~Visual examples of semantic maps predicted by SimIT compared with DRN. (c)~Quantitative segmentation results on the CholecSeg8k dataset. We report average F1 scores for six label classes over 213 test images over 213 test images. (d)~Quantitative segmentation results on the Cityscapes dataset over 500 test images.}
\label{seg_res}
\end{figure*}

For laparoscopy comparison, we report \emph{upper-bound} segmentation results from a ResNet50 network trained on annotated images from the CholecSeg8k dataset introduced by~\cite{hong2020cholecseg8k}, which is a subset of Cholec data~\cite{endonet}. 
In \Cref{seg_res}(a) a sample input image is shown together with its supervised ResNet50 segmentation as upper-bound; the semantic map predicted by SimIT used as a segmenter; and the ground-truth annotation for this test image. 
Average segmentation scores are reported in \Cref{seg_res}(c).
For the gaming application, we compare SimIT with the segmentation network DRN~\cite{yu2017dilated}, a standard technique for the Cityscapes dataset.
DRN was trained on the labelled training set, acting as a supervised upper-bound herein.
A qualitative sample comparison is shown in \Cref{seg_res}(b) with quantitative results tabulated in \Cref{seg_res}(d).
SimIT presents a fair performance despite not having seen any labeled real images and while not specifically targeting such segmentation problem.

For the ultrasound application, SimIT was trained with gray-scale label-maps, as this performed well for the main focus of label-to-image translation and also using one-hot label encoding (performed for the other two applications) was less stable in the parametrization of ultrasound training.
Without one-hot labels, our trained network fails to estimate meaningful label-maps from real ultrasound images.
This is mainly due to having nearly 80 different tissue classes and some classes with similar ultrasound texture and appearance, which makes the segmentation problem very challenging and, given no one-hot labels, also ill-posed as then a grayscale regression problem.

\section{Discussion}\label{sec12}
In this work we present a contrastive learning based unpaired image-label translation framework, by leveraging domain-specific simulations to generate photorealistic images from synthetic semantic label-maps.
We demonstrate the superior content-preservation performance of our proposed method across several datasets.
Our bidirectional framework as a by-product affords an image segmenter, which is demonstrated herein to provide approximate segmentations of real images.
Note that such segmentation by our method requires no annotations of real images in training, and it utilizes merely existing computational simulation outputs.
As demonstrated in \Cref{res_lapro,res_us,res_gta}(b), the unsupervised losses between label and image representations, \ie the cycle consistency loss~\cite{zhu2017unpaired} and the contrastive loss~\cite{park2020contrastive}, may lead to scene modification and \emph{semantic flipping}, \ie the content not being preserved semantically consistently.

To mitigate these, we leverage simulated images as intermediate image representation to bridge the gap between the source and target domains.
Among compared methods, ConPres is the closest to our work, as it also leverages simulated pairs to enforce content preservation.
To encourage the generator encoder to extract content-related features, ConPres uses a unified generator for three translation tasks: label-to-image, image-to-image, and image-to-label.
This, however, complicates the generator's task, leading to sub-optimal results in the label-to-image direction.
In comparison, we suggest to utilize task-specific generators, relaxing the constraints of each generator.
We accordingly leverage simulated images as surrogate targets only for loss computation, not as an additional generation task.
During our preliminary experiments, we found out that using pixel-level supervised losses, \eg L1/L2 loss, to assess content similarity between simulated and translated images is problematic due to the intrinsic appearance shift, lighting variations, and texture differences between two domains.
We also experimented with employing the discriminator as a feature extractor for computing feature-level losses, \eg the perceptual loss~\cite{johnson2016perceptual}, but the results were again not satisfactory.
In comparison to the above, the currently utilized patch-based contrastive loss is less affected by any appearance shift, and can therefore successfully discern and focus on content dissimilarity.
Together with our utilization of adaptive discriminator augmentation and limited discriminator receptive field, the contrasting of image features from the proposed addition of an image-to-label network has endowed results with substantial content preservation without degrading image realism.

The above-mentioned challenge of measuring image content similarity during training also reflects on the evaluation of inference results for structural preservation.
For the gaming application experiment, we have employed a pretrained segmentation network to assess the content difference between simulated and translated images.
This approach is only feasible when a large amount of annotated images from the target domain are available to train a segmentation network.
When such segmentation method is not available, choosing surrogate metrics for quantifying structural similarity is a non-trivial task.
Compared to metrics based on pixel-wise difference, SSIM is relatively less sensitive to appearance shifts.
Thus, we used SSIM to capture structural differences between laparoscopic images.
However, SSIM is less suitable for ultrasound images due to highly directional artifacts and the inherent speckle noise.
In ultrasound, bone surfaces are major anatomical landmarks which appear consistently as hyperechoic bands due to their relatively higher acoustic impedance~\cite{ozdemir2020delineating}.
We thus use bone surfaces extracted from simulated and translated ultrasound images using a phase-symmetry based method (described further in the Methods) for assessing structural preservation in ultrasound.

As noted from the qualitative results shown in \Cref{res_lapro,res_us,res_gta}, SimIT is not capable of recovering image features, which are not encoded in label-maps, such as lighting and depth information in laparoscopy and gaming applications, and the acoustic shadows in the ultrasound application.
Auxiliary scene information, \eg geometry and material information from the simulation, can be potentially integrated into SimIT as additional inputs, which can be a potential future direction to further improve content preservation.

Our proposed method can also have other uses.
For instance, in fields where annotated data is scarce and/or cannot be distributed due to privacy concerns, such as for medical image segmentation, there have been generative methods~\cite{kwon2019generation,pinaya2022brain} that produce image and label-map pairs based on random vectors, \eg to train supervised methods with large datasets.
In such a problem setting, our method can generate label-conditional images to establish a control on the generated dataset, which can help in, \eg creating balanced classes, removing biases in the original dataset, and emphasizing certain pathology.

\section{Methods}\label{sec11}
Herein we use the notation $X_y^z$ to represent the domain of any sample, where $X$ is the representation from \{$L$:label-map,\,$I$:image\}, and $y$ and $z$ are, respectively, the style (appearance) and content of such representation from \{$S$:simulated,\,$R$:real\}.
We aim to learn a generator $G:\mathbb{L^\text{S}_\text{S}}\mapsto\mathbb{I^\text{S}_\text{R}}$ which maps a simulated label-map $L_{\text{S}}^{\text{S}}\in \mathbb{L^\text{S}_\text{S}}$ to a real-appearing image $I^{\text{S}}_{\text{R}}\in \mathbb{I^\text{S}_\text{R}}$ while preserving the simulation-consistent semantic content, \ie $(\cdot)^\text{S}$.

\begin{figure*}
\centering
\includegraphics[width=1\textwidth]{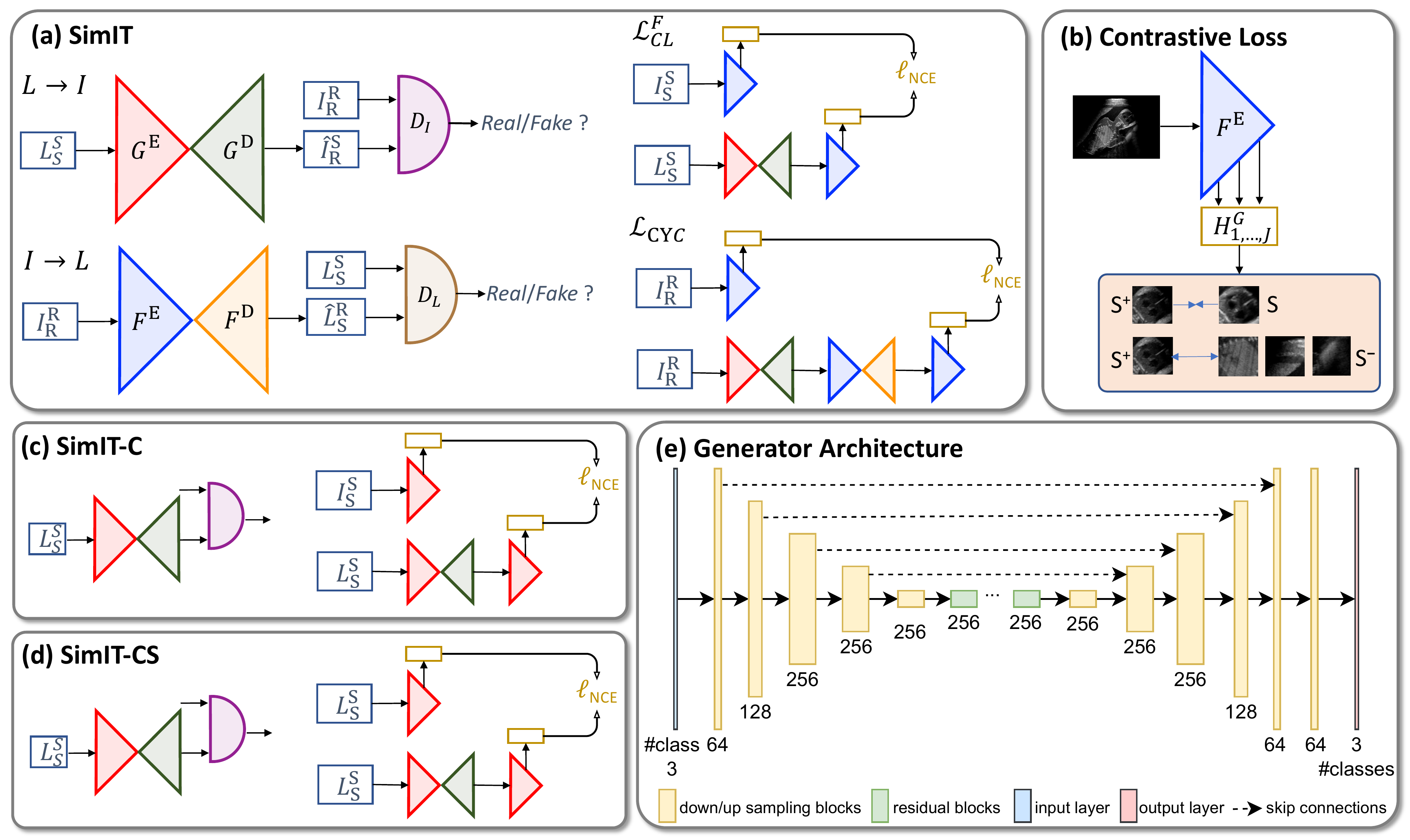}
\caption{(a)~Schematic overview of our proposed method SimIT with (b)~an illustration of contrastive loss. (c)~Schematic overview of the ablated version \mbox{SimIT-C}, and (d)~\mbox{SimIT-CS}. (e)~Schematics of the generator architecture. The number below each convolutional block indicates the channel number. For $G$ the input channel number is the number of classes (\#class) and the output channel number is the number of image channels (3 for RGB images), for $F$ vice versa.}
\label{method}
\end{figure*}

Generator $G$ is divided into an encoder $G^\mathrm{E}$ for extracting content-related features and a decoder $G^\mathrm{D}$ for generating target appearance.
It is possible to collect many real image examples $\{I_\text{R}^\text{R}\in\mathbb{I^\text{R}_\text{R}}\}$ and also to simulate label-image pairs $(L_\text{S}^\text{S},I_\text{S}^\text{S})$, but paired data of the intended source-target translation, \ie $(L_\text{S}^\text{S},I_\text{S}^\text{R})$, is inexistent and very challenging to procure.
The unpaired data described above does not allow for direct supervision in learning $G$.
Existing unpaired methods often change both content and style together, and the ones that aim content preservation only targets image-to-image translation, with methods we show herein not to simply extend to label-to-image translation targeted herein.
An overview of the methods can be followed in \Cref{method}.

\vspace{1ex}\noindent{\bf Generative adversarial training. }
For learning a generator $G$ and its discriminator $D_I$ \mbox{differentiating} images $I$ as real or fake, a non-saturating GAN loss with the R1 regularization~\cite{mescheder2018training} is used, \ie:
\begin{equation} 
    \mathcal{L}^\text{G}_\text{GAN}(\{L^\text{S}_\text{S}\}, \{I^\text{R}_\text{R}\}) =\mathbb{E}_{I} [\log (D_I(I^\text{R}_\text{R})-1)] 
    + \mathbb{E}_{L} [\log D_I(G(L^\text{S}_\text{S}))]
    + \frac{\gamma_{I}}{2}\mathbb{E}_{I} [\|\nabla D_I (I^\text{R}_\text{R})\|^2]
\end{equation}
with the regularization parameter $\gamma_{I}$.

\vspace{1ex}\noindent{\bf Label-to-image translation guided by simulation. }
Herein we propose to leverage information from simulations to achieve semantic preservation while learning $G$.
To that end, we utilize simulated (synthetic) images \mbox{$\{I_\text{S}^\text{S}\in\mathbb{I^\text{S}_\text{S}}\}$} during training, which can have paired input label-maps \mbox{$\{L_\text{S}^\text{S}\in\mathbb{L^\text{S}_\text{S}}\}$} generated from the existing simulation framework (\Cref{intro}(a)) and available for training.
We encourage scene-consistent translations using a contrastive loss~\cite{park2020contrastive} on image patches, where corresponding patches from the source and translated images (positive pairs) are brought closer in a learned feature space.
This space is defined as a projection of the manifold learned by the generator encoder, as illustrated in \Cref{method}(b).
Meanwhile non-corresponding (arbitrary) patches are treated as negative samples and hence pushed farther apart in that feature space.
Compared to the pixel-wise supervised losses, contrastive loss is known to be less affected by image appearance.
It was utilized in~\cite{park2020contrastive} for unpaired \emph{image-to-image} translation, \ie when both source and target are of the same representation being in image domain $\mathbb{I}$.
However, for \emph{label-to-image} translation, the source and target representations differ, \ie while each pixel in $L$ denotes a label, each pixel in $I$ denotes a color.
Thus, directly contrasting label and image features cannot successfully guide the network for the intended task; as also seen with the suboptimal performance of our ablation variant \mbox{SimIT-CS}.
To alleviate this problem, herein we leverage available simulated images $I^{\text{S}}_{\text{S}}\in\mathbb{I}^{\text{S}}_{\text{S}}$ as ``surrogate'' source images.
This implicitly enables the use of existing simulated images $I^\text{S}_\text{S})$.
Note that these images that require complex rendering operations are used for loss computation during the training of our method, so they are not needed during inference.
This is in contrast to the earlier works~\cite{tomar2021content,richter2022enhancing} where the rendered images are used as an input to the translation network and are thus complex rendering operations are still required during inference in real-time.

\vspace{1ex}\noindent{\bf Bidirectional label-image translation framework. }
To extract domain-consistent feature representations, we propose to employ an additional generator $F:\mathbb{I}_\text{R}^\text{R}\mapsto\mathbb{L}_\text{S}^\text{R}$ with $F(\cdot)=F^\text{D}(F^\text{E}(\cdot))$ consisting of an encoder $F^\text{E}$ and a decoder $F^\text{D}$, acting in the opposite direction for translating $I\rightarrow L$, \ie mapping an image back to a label-map.
Unlike~\cite{park2020contrastive} which contrasts features from $G^{\text{E}}$ operating on the source domain $\mathbb{L}$ with \emph{labels}, we propose to contrast the features of the segmenter encoder $F^{\text{E}}$ trained to extract features for inferring semantic content from images, and it is thus more suited for comparing \emph{image} similarity.

\vspace{1ex}\noindent{\bf Patch-based contrastive loss. }
For an input image $x$ and its translated image $\hat{x}=G^{\text{D}}(G^\text{E}(x))$, we contrast feature maps $z^{\text{F}}_j=H^{\text{G}}_j(F^{\text{E}}(x))$ and $\hat{z}^{\text{F}}_j=H^{\text{G}}_j(F^{\text{E}}(\hat{x}))$,
with a light-weight projection head $H_{j}^{\text{G}}$ mapping the features of $j$-th hidden layer in $F^{\text{E}}$ for training $G$.
Given $\hat{z}_j^{\text{F},s}$ as a \emph{query} feature at a given spatial location $s$ within the feature map $\hat{z}_j^{\text{F}}$ of the translated image, the corresponding input feature $z_j^{\text{F},s+}$ at the same location (denoted by +) is considered as a positive sample, while any other input feature $z_j^{\text{F},s-}$ at arbitrary locations act as negative samples.
Noise contrastive loss (NCE)~\cite{park2020contrastive} for the $j$-th layer feature can then be computed as the sum of contrastive loss for several $S_j$ at randomly sampled spatial locations as follows:
\begin{equation}
    \mathcal{L}_{\text{NCE},j}(z_j^{\text{F}}, \hat{z}_j^{\text{F}}) = \sum_{s=1}^{S_j} \mathcal{L}_{\text{CE}}(\hat{z}_j^{\text{F},s}, z_j^{\text{F},s+}, z_j^{\text{F},s-})
\end{equation}
with
\begin{equation}
    \mathcal{L}_\text{CE}(z, z^+, z^-) = \\ - \log \left[ \frac{\exp(d(z, z^+)/\tau)}{\exp(d(z, z^+)/\tau)+\sum_{z^-}\exp(d(z, z^-)/\tau)} \right],
\end{equation}
where $z$ is the feature vector of query; $z^+$ and $z^-$ are the feature vectors of positive and negative samples, respectively; $d(z_1, z_2)$ is a distance metric between two latent vectors (herein the cosine distance); and $\tau$ is a temperature parameter controlling the smoothing of joint likelihoods.

We herein propose to leverage the simulated image domain as surrogate source domain by computing contrastive loss between the translated image $\hat{I}_{R}^{S}=G^{\text{D}}(G^{\text{E}}(L_\text{S}^\text{S}))$ and simulated image $I_{S}^{S}$ paired to $L_{S}^{S}$, \ie:
\begin{equation}
    \mathcal{L}^\text{F}_\text{CL}\big(\{L^\text{S}_\text{S}\}\big)=\mathbb{E}_{L}\sum_{j=1}^{J} \mathcal{L}_{\text{NCE},j}\Big(H^{\text{G}}_j\big(F^{\text{E}}(I_\text{S}^\text{S})\big), H^{\text{G}}_j\big(F^{\text{E}}(\hat{I}_\text{R}^\text{S})\big)\Big)
\end{equation}
computed over $J$ feature layers to contrast information at different resolutions.

For a more expressive and distinctive label space, we herein encode label-maps as one-hot representations, which prevents misinterpretation of categorical labels suitable for class separation as pixel intensities for regression.

\vspace{1ex}\noindent{\bf Learning image-to-label translation using cycle consistency loss. }
In this work, we treat and train the image-to-label translator $F$ to perform pixel-wise semantic labeling task, known as image segmentation.
Based on our cyclic framework, we use the existing label-to-image mapping network $G$ to assess such segmentation accuracy, hence also obviating a need for pixel-wise annotations of real images, which are difficult to procure.
With that, we compute a cycle reconstruction loss between a real image $I^\text{R}_\text{R}$ and the image reconstructed from the predicted label-map $\tilde{I}^\text{R}_\text{R} = G(F(I^\text{R}_\text{R}))$.
Image similarity is measured using the NCE loss, as it is less sensitive to appearance shifts, as follows
\begin{equation}
    \mathcal{L}_\text{CYC}\big(\{I^\text{R}_\text{R}\}\big)=\mathbb{E}_{L}\sum_{j=1}^{J} \mathcal{L}_{\text{NCE},j}\Big(H^{\text{F}}_j\big(F^{\text{E}}(I_\text{R}^\text{R})\big), H^{\text{F}}_j\big(F^{\text{E}}(\tilde{I}_\text{R}^\text{R})\big)\Big)
\end{equation}
with the projection head $H_j^{F}$ for the $j$-th layer feature of encoder $F$.
For $F$ and its discriminator $D_L$ for the label representation direction, we employ a GAN training objective $\mathcal{L}^\text{F}_\text{GAN}(\{I^\text{R}_\text{R}\}, \{L^\text{S}_\text{S}\})$ similar to the original direction $G$, but with a different regularization parameter $\gamma_{L}$.

\vspace{1ex}\noindent{\bf Training objective. }
A schematic illustration of the proposed method SimIT summarizing the above components is shown in \Cref{method}(a).
Network training is performed by alternately optimizing the following two losses:
\begin{align}
    \mathcal{L}_{\text{G}}(\{L^\text{S}_\text{S}\}, \{I^\text{R}_\text{R}\}) &= \mathcal{L}^\text{G}_\text{GAN}(L^\text{S}_\text{S}, I^\text{R}_\text{R}) + \lambda_G \cdot  \mathcal{L}^\text{F}_\text{CL}(L^\text{S}_\text{S})\\
    \mathcal{L}_{\text{F}}(\{L^\text{S}_\text{S}\}, \{I^\text{R}_\text{R}\}) &= \mathcal{L}^\text{F}_\text{GAN}(L^\text{S}_\text{S}, I^\text{R}_\text{R}) + \lambda_F \cdot  \mathcal{L}_\text{CYC}(I^\text{R}_\text{R})
\end{align}
with the loss weighting parameters $ \lambda_G$ and $\lambda_F$.

We compare our full model against the ablated variant \mbox{SimIT-C} by excluding the inverse mapping $\mathcal{L}_{\text{F}}$ with the cyclic loss, as seen in \Cref{method}(c).
Further ablating the paired simulation images yield the variant \mbox{SimIT-CS} that instead uses the labels for contrasting (\Cref{method}(d)).
For both ablated variants, encoder $G^{\text{E}}$ is used for computing the contrastive loss $\mathcal{L}^\text{G}_\text{CL}$.

\vspace{1ex}\noindent{\bf Network architecture. }
We build our method on the StyleGAN2 framework~\cite{karras2020analyzing} for adversarial training.
We accordingly use a ResNet-based generator architecture~\cite{park2020contrastive} with four down- and up-sampling layers and 6 residual blocks (\Cref{method}(d)).
We use skip connections between the down- and upsampling layers to avoid information loss.
For the image synthesis decoder $G^{\text{E}}$ we use weight demodulation~\cite{karras2020analyzing}
\begin{equation}
    w''_{ijk} = \frac{w'_{ijk}}{\sqrt{\sum_{i,k} w'_{ijk^2} + \epsilon}} \quad \text{with} \quad w'_{ijk} = s_i \cdot w_{ijk}
\end{equation}
where $w_{ijk}$ is the convolution weight from the $i$-th input feature map to the $j$-th output feature map; $k$ denotes the spatial footprint of the convolution; and the multiplier
$s_i$ is set to 1.
To provide stochastic means for texture synthesis, especially important to generate the noisy speckle patterns of ultrasound images, we perturb each feature layer with an additive Gaussian (noise) image scaled by learned weights following~\cite{karras2019style}.
The output layer for $G$ and $F$ is linear and sigmoid, respectively.
We use ReLU activation for all intermediate layers.
For both $D_I$ and $D_L$, we adopt the feedforward discriminator architecture in~\cite{karras2020analyzing}.
In training we use randomly cropped image patches, which enables the discriminator to ignore global scene differences between simulation and real domains.

\vspace{1ex}\noindent{\bf Experimental data utilization. }
\emph{LaparoSim} consists of 1850 synthetic laparoscopic image-label pairs simulated from a 3D abdominal model.
We randomly split this data into train-validation-test sets with 80-10-10\% ratio.
\emph{Style-C} consist of 2100 images from the Cholec dataset.
We excluded all frames with surgical tools, since surgical tools are not handled in our simulation.
For Style-C testing, we used the 213 frames that has ground-truth labels provided in~\cite{hong2020cholecseg8k}, and the remaining frames were used as Style-C training data.
\emph{Style-H} consists of 2262 frames in total, which was randomly split in 80-20\% ratio, respectively, for training and testing.
Some Style-H images had major blurring artifacts due to camera motion, so we manually removed any blurry frames, since we treat the frames separately without any temporal information and such temporal effect is also not represented in the label-maps.
All the images were resized to $256\times432$.

\emph{USSim} consists of $6669$ simulated image-label pairs, which we resized to $256\times 354$ and randomly split into training-validation-test sets with 80-10-10\% ratio.
\emph{GE-E10 style} consists of 2328 ultrasound images from 24 patients.
We randomly selected images from 20 patients for training and 4 for testing, resulting in 1902 training images and 426 test images.

\emph{GTA} dataset~\cite{richter2016playing} contains 24\,966 image-label pairs.
We followed its original train-validation-test split.
\emph{Cityscapes} dataset~\cite{Cordts2016Cityscapes} contains 3475 image-label pairs of street scenes from German cities.
We used its original training set for our network training, and its validation set with ground-truth labels for testing our semantic segmentation output.
As in~\cite{park2020contrastive}, we resized all the images to $256\times256$.

\vspace{1ex}\noindent{\bf Implementation. }
We implemented our method in PyTorch~\cite{paszke2019pytorch}.
We used Adam~\cite{kingma2014adam} optimizer with parameters $\beta=(0, 0.99)$ and a learning rate of $10^{-3}$ for $G$ and $10^{-4}$ for $F$.
We applied adaptive discriminator augmentation using its default hyperparameters~\cite{karras2020training}.
The generator is trained on image patches of size $256\times256$ while the discriminator receptive field is $64\times64$.
Our network training involves alternating updates of $G$ and $F$.
We trained our models for 400 epochs.
To compute the contrastive loss, we extract features from the four (stride-2) convolution layers in the encoder at 256 locations randomly selected for each mini-batch.
We use a two-layer MLP with 256 units at each layer and ReLU activation for $H^{\text{G}}$, and the identity mapping for $H^{\text{F}}$.
NCE temperature parameter $\tau$ is set to 0.07 following~\cite{park2020contrastive}.
Generator loss weighting $\lambda_G$ is set to 5 for all the experiments.
$\lambda_F$ is set to 1 for the laparoscopy and ultrasound, and 0.5 for the gaming experiment.
R1 regularization parameters $\gamma_{I}$ and $\gamma_{L}$ are set to 0.01 and 1.0, respectively, for all the experiments.
For all compared methods we used their public implementations provided by the corresponding authors with their default hyperparameters.

\vspace{1ex}\noindent{\bf Evaluation metrics. }
We use the following for quantitative evaluation:
\vspace{-0.5ex}\begin{itemize}[labelwidth=-2ex,leftmargin=0pt]
\item[$\bullet$] {\bf Image realism.}
\emph{Fr\'{e}chet inception distance} (FID)~\cite{heusel2017gans} is common for assessing the quality of images generated by GANs, by comparing the feature distribution between two sets of images, herein real and translated, using feature vectors of an ImageNet-pretrained Inception network.
\emph{Kernel inception distance} (KID)~\cite{binkowski2018demystifying} is an alternative metric to evaluate GAN performance.
KID is computed as the squared maximum mean-discrepancy between the features of Inception network.
KID is then not biased by the number of samples used, unlike FID.
\item[$\bullet$] {\bf Content preservation.}
For laparoscopy images, content preservation is assessed using structural similarity between simulated and translated images, quantified via 
\emph{Structural similarity index} (SSIM) computed as $\textrm{SSIM}(x, y)$=$\frac{(2\mu_x\mu_y+c_1)(2\sigma_{xy}+c_2)}{(\mu_x^2+\mu_y^2+c_1)(\sigma_x^2+\sigma_y^2+c_2)}$ with regularization constants $c_1$ and $c_2$, local intensity means $\mu_x$ and $\mu_y$, local standard deviations $\sigma_x$ and $\sigma_y$, and $\sigma_{xy}$ being covariance between $x$ and $y$.
To compute this metric, we used the python package \emph{scikit-image} with its default parameters.
For ultrasound images, due to potential artifacts, typical speckle noise, and a lack of sharp edges, we instead used the similarity of bone surfaces for assessing structure preservation.
To that end, we extracted bone surfaces from each image using~\cite{hacihaliloglu2009bone}.
This method is based on local phase symmetry in B-mode images, and operates by aggregating images filtered by log-Gabor kernels with different orientations $r$ and scales $m$ defined as
\begin{equation}
    G_{r,m}(\omega, \phi) = \exp \left(-\frac{\log(\omega/\omega_0)^2}{2\log(\kappa_m/\omega_0)^2} - \frac{(\phi-\phi_r)^2}{2\sigma_\phi^2}\right),
\end{equation}
where parameters $\phi_r$, $\omega_0$, $\kappa_m$, and $\sigma_\phi$ define the filter orientation, center frequency, scaling factor, and angular bandwidth of the employed filters, respectively.
Following~\cite{ozdemir2020delineating}, we set $\kappa_m/\omega_0 = 0.25$ and $\phi_r=[\frac{1}{6}\pi, \frac{3}{6}\pi, \frac{5}{6}\pi]$.
To assess preservation, we report \emph{intersection over union} (IoU) of pixels belonging to bone surfaces extracted from corresponding simulated and translated images.
We exclude from computations the top 25 pixels of images corresponding to skin reflections.
\item[$\bullet$] {\bf Segmentation.} 
For the laparoscopic CholecSeg8k dataset, we trained a semantic segmentation network with ResNet50 architecture initialized with weights pretrained on the ImageNet using the pytorch segmentation library~\cite{Iakubovskii:2019}, following the training settings from a public implementation on the Kaggle repository of this dataset.
We randomly picked video24 for validation, video\{09,17,26,28,43\} for testing, and the rest for training.
We report F1 score for six classes that are also in our simulation.
For the Cityscapes dataset in the gaming application, we trained a segmentation network suggested for this dataset in~\cite{park2020contrastive}, with the DRN-D22 architecture~\cite{yu2017dilated} at $256\times128$ resolution with the default parameters from its public implementation.
Following~\cite{park2020contrastive}, we report Cityscapes semantic segmentation results using \emph{mean average precision} (mAP) over the classes; \emph{pixel-wise accuracy} (pixAcc) as the percentage of correctly classified pixels; and \emph{average class accuracy} (classAcc) over given classes.

\end{itemize}

\section*{Appendix}
\begin{figure*}
\centering
\includegraphics[width=1\textwidth]{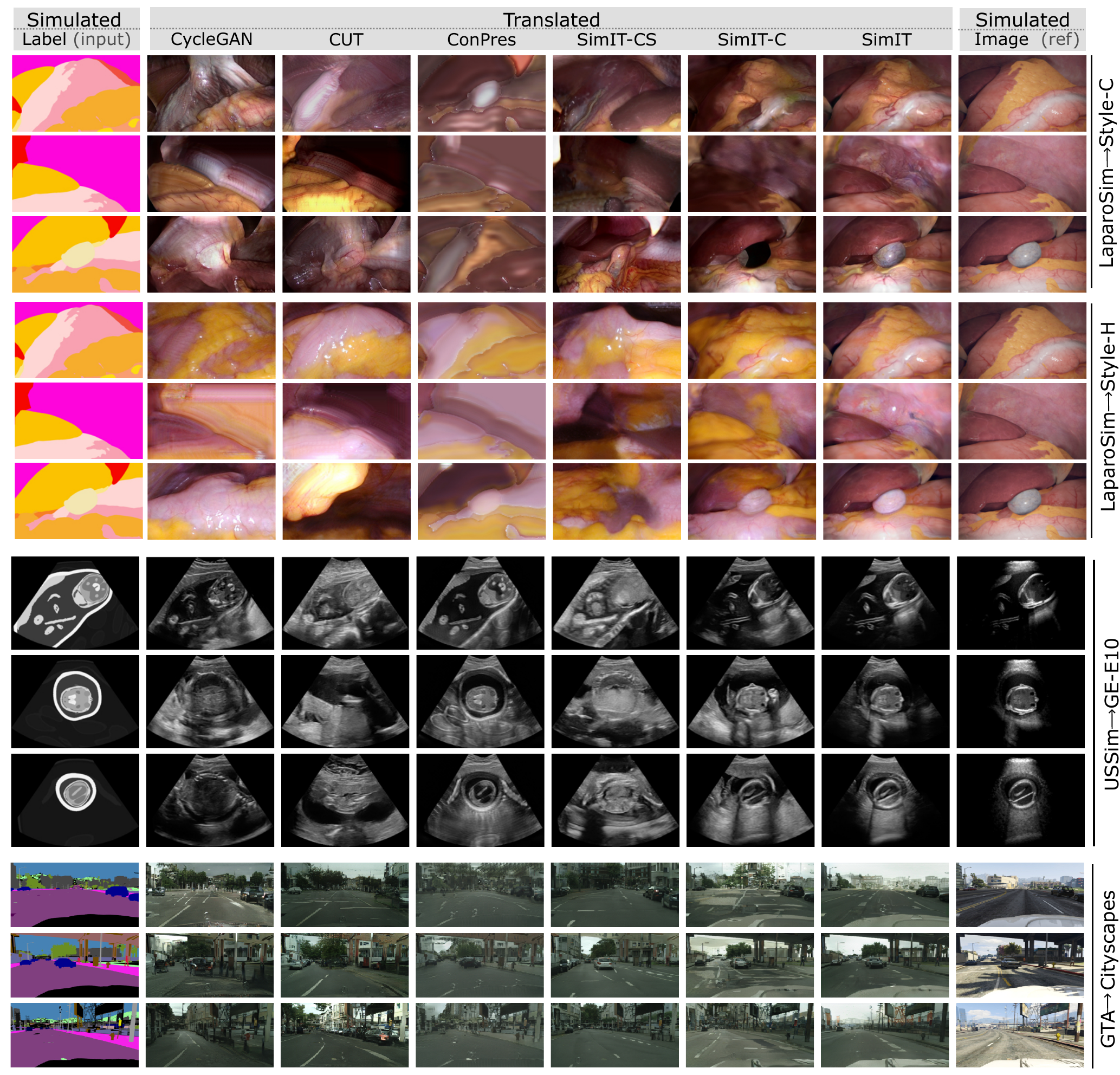}
\caption{Additional qualitative label-to-image translation results for the laparoscopic simulation application Style-C (top three rows) and Style-H (next three rows), the ultrasound simulation application (next thre rows), and the gaming application (last three rows).}
\label{supp1}
\end{figure*}

\begin{figure*}
\centering
\includegraphics[width=1\textwidth]{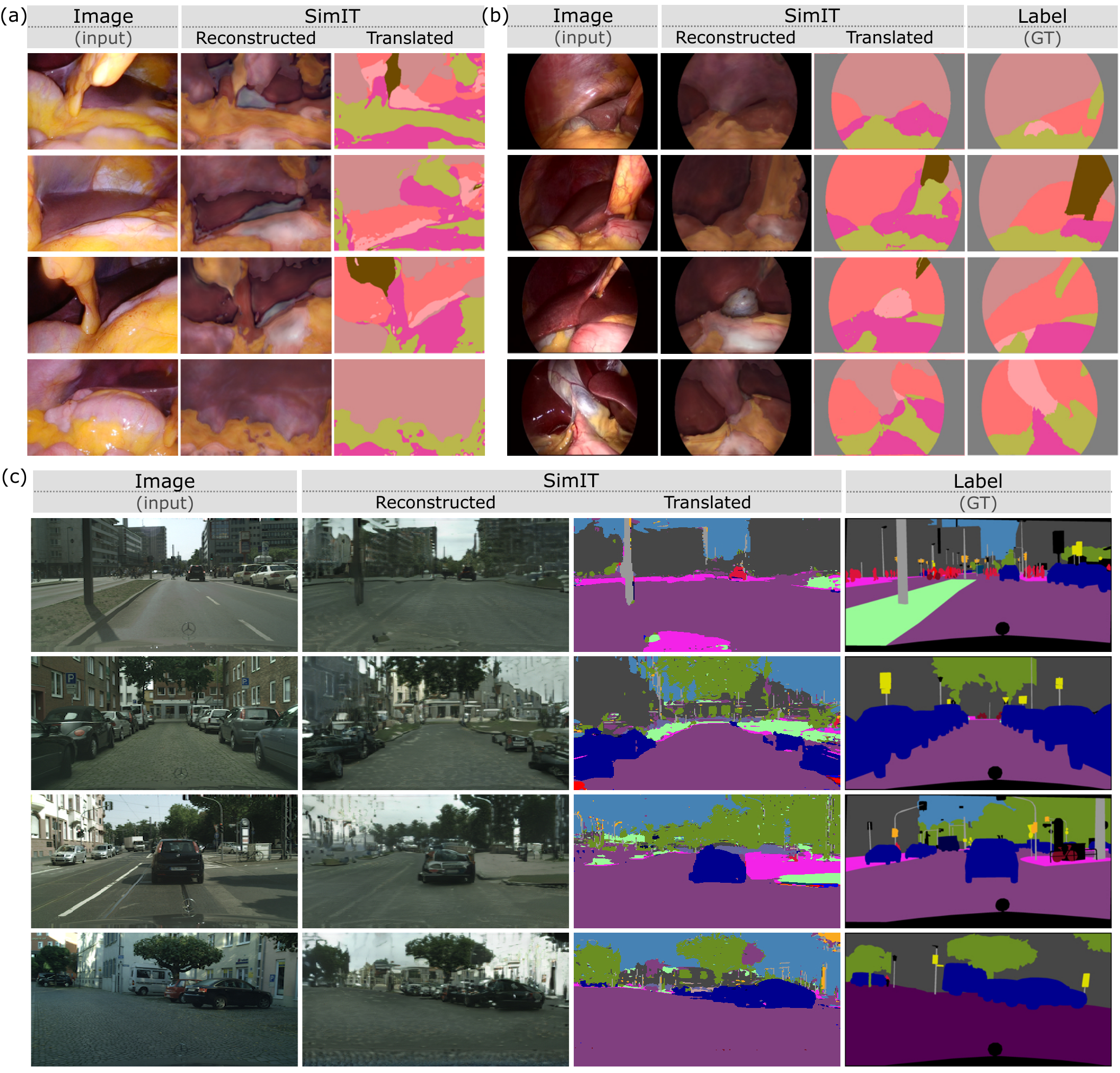}
\caption{Additional image-to-label translation results of SimIT with the input real image $I_\text{R}^\text{R}$, translated result $F(I_\text{R}^\text{R})$, reconstructed image $G(F(I_\text{R}^\text{R}))$, and ground truth label-maps for reference; for (a)~in-house laparoscopic, (b)~CholecSeg8k, and (c)~Cityscapes images.
No ground-truth annotation is available for the dataset in (a).
Note that although the reconstructed images are generated from translated images, they are ordered in-reverse to place the same representations with the input and GT next to each other for easier visual comparison.}
\label{supp2}
\end{figure*}

Additional sample images of image-to-label and label-to-image translations are shown in \Cref{supp1,supp2}, respectively.

\bibliographystyle{IEEEtran}
\bibliography{ref}

\end{document}